\documentclass[conference]{IEEEtran}
\IEEEoverridecommandlockouts

\usepackage{cite}
\usepackage{amsmath,amssymb,amsfonts}
\usepackage{algorithm}
\usepackage{graphicx}
\usepackage{textcomp}
\usepackage{xcolor}
\usepackage{algpseudocode}
\usepackage{float}
\usepackage{subcaption}
\usepackage{adjustbox}

\usepackage{acronym}

\graphicspath{{images/}} 

\begin{document}

\acrodef{ADC}[ADC]{Analog to Digital Converter}
\acrodef{ADEXP}[AdExp-I\&F]{Adaptive-Exponential Integrate and Fire}
\acrodef{ADM}[ADM]{Asynchronous Delta Modulator}
\acrodef{AER}[AER]{Address-Event Representation}
\acrodef{AEX}[AEX]{AER EXtension board}
\acrodef{AE}[AE]{Address-Event}
\acrodef{AFM}[AFM]{Atomic Force Microscope}
\acrodef{AGC}[AGC]{Automatic Gain Control}
\acrodef{AI}[AI]{Artificial Intelligence}
\acrodef{AMDA}[AMDA]{AER Motherboard with D/A converters}
\acrodef{ANN}[ANN]{Artificial Neural Network}
\acrodef{API}[API]{Application Programming Interface}
\acrodef{APMOM}[APMOM]{Alternate Polarity Metal On Metal}
\acrodef{ARM}[ARM]{Advanced RISC Machine}
\acrodef{ASIC}[ASIC]{Application Specific Integrated Circuit}
\acrodef{AdExp}[AdExp-IF]{Adaptive Exponential Integrate-and-Fire}
\acrodef{BCM}[BMC]{Bienenstock-Cooper-Munro}
\acrodef{BD}[BD]{Bundled Data}
\acrodef{BEOL}[BEOL]{Back-end of Line}
\acrodef{BG}[BG]{Bias Generator}
\acrodef{BMI}[BMI]{Brain-Machince Interface}
\acrodef{BTB}[BTB]{band-to-band tunnelling}
\acrodef{CAD}[CAD]{Computer Aided Design}
\acrodef{CAM}[CAM]{Content Addressable Memory}
\acrodef{CAVIAR}[CAVIAR]{Convolution AER Vision Architecture for Real-Time}
\acrodef{CA}[CA]{Cortical Automaton}
\acrodef{CCN}[CCN]{Cooperative and Competitive Network}
\acrodef{CDR}[CDR]{Clock-Data Recovery}
\acrodef{CFC}[CFC]{Current to Frequency Converter}
\acrodef{CHP}[CHP]{Communicating Hardware Processes}
\acrodef{CMIM}[CMIM]{Metal-insulator-metal Capacitor}
\acrodef{CML}[CML]{Current Mode Logic}
\acrodef{CMOL}[CMOL]{Hybrid CMOS nanoelectronic circuits}
\acrodef{CMOS}[CMOS]{Complementary Metal-Oxide-Semiconductor}
\acrodef{CNN}[CNN]{Convolutional Neural Network}
\acrodef{COTS}[COTS]{Commercial Off-The-Shelf}
\acrodef{CPG}[CPG]{Central Pattern Generator}
\acrodef{CPLD}[CPLD]{Complex Programmable Logic Device}
\acrodef{CPU}[CPU]{Central Processing Unit}
\acrodef{CSM}[CSM]{Cortical State Machine}
\acrodef{CSP}[CSP]{Constraint Satisfaction Problem}
\acrodef{CTXCTL}[CTXCTL]{CortexControl}
\acrodef{CV}[CV]{Coefficient of Variation}
\acrodef{DAC}[DAC]{Digital to Analog Converter}
\acrodef{DAS}[DAS]{Dynamic Auditory Sensor}
\acrodef{DAVIS}[DAVIS]{Dynamic and Active Pixel Vision Sensor}
\acrodef{DBN}[DBN]{Deep Belief Network}
\acrodef{DFA}[DFA]{Deterministic Finite Automaton}
\acrodef{DIBL}[DIBL]{drain-induced-barrier-lowering}
\acrodef{DI}[DI]{delay insensitive}
\acrodef{DMA}[DMA]{Direct Memory Access}
\acrodef{DNF}[DNF]{Dynamic Neural Field}
\acrodef{DNN}[DNN]{Deep Neural Network}
\acrodef{DOF}[DOF]{Degrees of Freedom}
\acrodef{DPE}[DPE]{Dynamic Parameter Estimation}
\acrodef{DPI}[DPI]{Differential Pair Integrator}
\acrodef{DRAM}[DRAM]{Dynamic Random Access Memory}
\acrodef{DRRZ}[DR-RZ]{Dual-Rail Return-to-Zero}
\acrodef{DR}[DR]{Dual Rail}
\acrodef{DSP}[DSP]{Digital Signal Processor}
\acrodef{DVS}[DVS]{Dynamic Vision Sensor}
\acrodef{DYNAP}[DYNAP]{Dynamic Neuromorphic Asynchronous Processor}
\acrodef{EBL}[EBL]{Electron Beam Lithography}
\acrodef{EDVAC}[EDVAC]{Electronic Discrete Variable Automatic Computer}
\acrodef{EEG}[EEG]{electroencephalography}
\acrodef{EIN}[EIN]{Excitatory-Inhibitory Network}
\acrodef{EM}[EM]{Expectation Maximization}
\acrodef{EPSC}[EPSC]{Excitatory Post-Synaptic Current}
\acrodef{EPSP}[EPSP]{Excitatory Post-Synaptic Potential}
\acrodef{EZ}[EZ]{Epileptogenic Zone}
\acrodef{FDSOI}[FDSOI]{Fully-Depleted Silicon on Insulator}
\acrodef{FET}[FET]{Field-Effect Transistor}
\acrodef{FFT}[FFT]{Fast Fourier Transform}
\acrodef{FI}[F-I]{Frequency-Current}
\acrodef{FPGA}[FPGA]{Field Programmable Gate Array}
\acrodef{FR}[FR]{Fast Ripple}
\acrodef{FSA}[FSA]{Finite State Automaton}
\acrodef{FSM}[FSM]{Finite State Machine}
\acrodef{GIDL}[GIDL]{gate-induced-drain-leakage}
\acrodef{GOPS}[GOPS]{Giga-Operations per Second}
\acrodef{GPU}[GPU]{Graphical Processing Unit}
\acrodef{GT}[GT]{Ground Truth}
\acrodef{GUI}[GUI]{Graphical User Interface}
\acrodef{HAL}[HAL]{Hardware Abstraction Layer}
\acrodef{HFO}[HFO]{High Frequency Oscillation}
\acrodef{HH}[H\&H]{Hodgkin \& Huxley}
\acrodef{HMM}[HMM]{Hidden Markov Model}
\acrodef{HRS}[HRS]{High-Resistive State}
\acrodef{HR}[HR]{Human Readable}
\acrodef{HSE}[HSE]{Handshaking Expansion}
\acrodef{HW}[HW]{Hardware}
\acrodef{ICT}[ICT]{Information and Communication Technology}
\acrodef{IC}[IC]{Integrated Circuit}
\acrodef{IEEG}[iEEG]{intracranial electroencephalography}
\acrodef{IF2DWTA}[IF2DWTA]{Integrate \& Fire 2--Dimensional WTA}
\acrodef{IFSLWTA}[IFSLWTA]{Integrate \& Fire Stop Learning WTA}
\acrodef{IF}[I\&F]{Integrate-and-Fire}
\acrodef{IMU}[IMU]{Inertial Measurement Unit}
\acrodef{INCF}[INCF]{International Neuroinformatics Coordinating Facility}
\acrodef{INI}[INI]{Institute of Neuroinformatics}
\acrodef{IO}[I/O]{Input/Output}
\acrodef{IPSC}[IPSC]{Inhibitory Post-Synaptic Current}
\acrodef{IPSP}[IPSP]{Inhibitory Post-Synaptic Potential}
\acrodef{IP}[IP]{Intellectual Property}
\acrodef{ISI}[ISI]{Inter-Spike Interval}
\acrodef{IoT}[IoT]{Internet of Things}
\acrodef{JFLAP}[JFLAP]{Java - Formal Languages and Automata Package}
\acrodef{LEDR}[LEDR]{Level-Encoded Dual-Rail}
\acrodef{LFP}[LFP]{Local Field Potential}
\acrodef{LLC}[LLC]{Low Leakage Cell}
\acrodef{LNA}[LNA]{Low-Noise Amplifier}
\acrodef{LPF}[LPF]{Low Pass Filter}
\acrodef{LRS}[LRS]{Low-Resistive State}
\acrodef{LSM}[LSM]{Liquid State Machine}
\acrodef{LTD}[LTD]{Long Term Depression}
\acrodef{LTI}[LTI]{Linear Time-Invariant}
\acrodef{LTP}[LTP]{Long Term Potentiation}
\acrodef{LTU}[LTU]{Linear Threshold Unit}
\acrodef{LUT}[LUT]{Look-Up Table}
\acrodef{LVDS}[LVDS]{Low Voltage Differential Signaling}
\acrodef{MCMC}[MCMC]{Markov-Chain Monte Carlo}
\acrodef{MEMS}[MEMS]{Micro Electro Mechanical System}
\acrodef{MFR}[MFR]{Mean Firing Rate}
\acrodef{MIM}[MIM]{Metal Insulator Metal}
\acrodef{MLP}[MLP]{Multilayer Perceptron}
\acrodef{MOSCAP}[MOSCAP]{Metal Oxide Semiconductor Capacitor}
\acrodef{MOSFET}[MOSFET]{Metal Oxide Semiconductor Field-Effect Transistor}
\acrodef{MOS}[MOS]{Metal Oxide Semiconductor}
\acrodef{MRI}[MRI]{Magnetic Resonance Imaging}
\acrodef{NCS}[NCS]{Neuromorphic Cognitive Systems}
\acrodef{NDFSM}[NDFSM]{Non-deterministic Finite State Machine} 
\acrodef{ND}[ND]{Noise-Driven}
\acrodef{NEF}[NEF]{Neural Engineering Framework}
\acrodef{NHML}[NHML]{Neuromorphic Hardware Mark-up Language}
\acrodef{NIL}[NIL]{Nano-Imprint Lithography}
\acrodef{NMDA}[NMDA]{N-Methyl-D-Aspartate}
\acrodef{NME}[NE]{Neuromorphic Engineering}
\acrodef{NN}[NN]{Neural Network}
\acrodef{NOC}[NoC]{Network-on-Chip}
\acrodef{NRZ}[NRZ]{Non-Return-to-Zero}
\acrodef{NSM}[NSM]{Neural State Machine}
\acrodef{OR}[OR]{Operating Room}
\acrodef{OTA}[OTA]{Operational Transconductance Amplifier}
\acrodef{PCB}[PCB]{Printed Circuit Board}
\acrodef{PCHB}[PCHB]{Pre-Charge Half-Buffer}
\acrodef{PCM}[PCM]{Phase Change Memory}
\acrodef{PE}[PE]{Phase Encoding}
\acrodef{PFA}[PFA]{Probabilistic Finite Automaton}
\acrodef{PFC}[PFC]{prefrontal cortex}
\acrodef{PFM}[PFM]{Pulse Frequency Modulation}
\acrodef{PR}[PR]{Production Rule}
\acrodef{PSC}[PSC]{Post-Synaptic Current}
\acrodef{PSP}[PSP]{Post-Synaptic Potential}
\acrodef{PSTH}[PSTH]{Peri-Stimulus Time Histogram}
\acrodef{QDI}[QDI]{Quasi Delay Insensitive}
\acrodef{RAM}[RAM]{Random Access Memory}
\acrodef{RA}[RA]{Resected Area}
\acrodef{RDF}[RDF]{random dopant fluctuation}
\acrodef{RELU}[ReLu]{Rectified Linear Unit}
\acrodef{RLS}[RLS]{Recursive Least-Squares}
\acrodef{RMSE}[RMSE]{Root Mean Squared-Error}
\acrodef{RMS}[RMS]{Root Mean Squared}
\acrodef{RNN}[RNN]{Recurrent Neural Networks}
\acrodef{ROLLS}[ROLLS]{Reconfigurable On-Line Learning Spiking}
\acrodef{RRAM}[R-RAM]{Resistive Random Access Memory}
\acrodef{R}[R]{Ripples}
\acrodef{SAC}[SAC]{Selective Attention Chip}
\acrodef{SAT}[SAT]{Boolean Satisfiability Problem}
\acrodef{SCX}[SCX]{Silicon CorteX}
\acrodef{SD}[SD]{Signal-Driven}
\acrodef{SEM}[SEM]{Spike-based Expectation Maximization}
\acrodef{SLAM}[SLAM]{Simultaneous Localization and Mapping}
\acrodef{SNN}[SNN]{Spiking Neural Network}
\acrodef{SNR}[SNR]{Signal to Noise Ratio}
\acrodef{SOC}[SOC]{System-On-Chip}
\acrodef{SOI}[SOI]{Silicon on Insulator}
\acrodef{SOZ}[SOZ]{Seizure Onset Zone}
\acrodef{SP}[SP]{Separation Property}
\acrodef{SRAM}[SRAM]{Static Random Access Memory}
\acrodef{STDP}[STDP]{Spike-Timing Dependent Plasticity}
\acrodef{STD}[STD]{Short-Term Depression}
\acrodef{STP}[STP]{Short-Term Plasticity}
\acrodef{STT-MRAM}[STT-MRAM]{Spin-Transfer Torque Magnetic Random Access Memory}
\acrodef{STT}[STT]{Spin-Transfer Torque}
\acrodef{SW}[SW]{Software}
\acrodef{TCAM}[TCAM]{Ternary Content-Addressable Memory}
\acrodef{TFT}[TFT]{Thin Film Transistor}
\acrodef{TLE}[TLE]{Temporal Lobe Epilepsy}
\acrodef{USB}[USB]{Universal Serial Bus}
\acrodef{VHDL}[VHDL]{VHSIC Hardware Description Language}
\acrodef{VLSI}[VLSI]{Very Large Scale Integration}
\acrodef{VOR}[VOR]{Vestibulo-Ocular Reflex}
\acrodef{WCST}[WCST]{Wisconsin Card Sorting Test}
\acrodef{WTA}[WTA]{Winner-Take-All}
\acrodef{XML}[XML]{eXtensible Mark-up Language}
\acrodef{divmod3}[DIVMOD3]{divisibility of a number by three}
\acrodef{hWTA}[hWTA]{hard Winner-Take-All}
\acrodef{sWTA}[sWTA]{soft Winner-Take-All}

\title{Cortical-inspired placement and routing: minimizing the memory resources in multi-core neuromorphic processors}

\author{\IEEEauthorblockN{
Vanessa R. C. Leite,
Zhe Su,
Adrian M. Whatley,
Giacomo Indiveri}
\IEEEauthorblockA{
Institute of Neuroinformatics,
University of Zurich and ETH Zurich\\
Email: vanessa@ini.uzh.ch}

\thanks{This work was partially supported by the European Research Council (ERC) under the European Union’s Horizon 2020 Research and Innovation Program Grant Agreement No. 724295 (NeuroAgents), and by the Electronic Component Systems for European Leadership (ECSEL) joint undertaking Grant Agreement No. 876925 (ANDANTE)}

}

\maketitle

\begin{abstract}
  Brain-inspired event-based neuromorphic processing systems have been emerging as a promising technology in particular for bio-medical circuits and systems.
  However, both neuromorphic and biological implementations of neural networks have critical energy and memory constraints.
  To minimize the use of memory resources in multi-core neuromorphic processors, we propose a network design approach that takes inspiration from biological neural networks.
  We use this approach to design a new routing scheme optimized for small-world networks and, at the same time, to present a hardware-aware placement algorithm that optimizes the allocation of resources for small-world network models.
  We validate the algorithm with a canonical small-world network and present preliminary results for other networks derived from it.
\end{abstract}


\section{Introduction}
\label{sec:introduction}
The large energy costs of \ac{DNN} and \ac{AI} algorithms are pushing the development of domain-specific hardware accelerators~\cite{Ibtesam_etal21}.
Neuromorphic processors are a class of \ac{AI} hardware accelerators that implement computational models of \acp{SNN} adopting in-memory computing strategies and brain-inspired principles of computation~\cite{Roy_etal19,Chicca_Indiveri20,Sebastian_etal20}.
They represent a very promising approach, especially for edge-computing and bio-signal processing applications, as they have the potential to reduce power consumption to ultra-low (e.g., sub-milliwatt) figures.
However, the requirement of \ac{SNN} hardware accelerators to store the state of each neuron, combined with their in-memory computing circuit design techniques leads to very large area consumption figures, which limits the sizes and numbers of parameters of the networks that they can implement.

The current strategy used to support the integration of large \ac{SNN} models in these accelerators is to use multi-core architectures~\cite{Moradi_etal18,Merolla_etal14a,Akopyan_etal15,Davies_etal18,Furber_Bogdan20}.
In these architectures, each core either \emph{emulates} with analog circuits~\cite{Moradi_etal18} or \emph{simulates} with time-multiplexed digital circuits~\cite{Akopyan_etal15,Davies_etal18,Furber_Bogdan20} neuro-synaptic arrays in which both the synaptic weight matrix and the network connectivity routing memory blocks occupy a significant proportion of the total layout area.
Although the advent of nano-scale memristive devices can mitigate this problem by enabling the construction of dense cross-bar array structures for storing the weight matrices~\cite{Sebastian_etal20}, the problem of allocating routing and connectivity resources to allow arbitrary networks at scale is of a fundamental nature that even memristors or 3D-VLSI technologies cannot solve~\cite{Laughlin_Sejnowski03}.

Finding trade-offs to optimize both weight-matrix and connectivity/routing memory structures in multi-core neuromorphic processors can therefore have a significant impact on their total chip die area and on the size of the networks they can implement.
Following the original neuromorphic engineering approach~\cite{Mead90}, in this paper, we look at animal brains for inspiration and propose brain-inspired architectures and strategies to reduce the memory needed to place and route networks, thus reducing the total chip die area.

Specifically, we show that, by focusing on small-world network connectivity, we can implement trade-offs that minimize memory consumption requirements while still enabling the design of \ac{SNN} architectures that can solve a wide range of relevant ``edge-computing'' problems, i.e., the types of sensory-motor processing problems that animals must solve in the real world.

\section{Neural network connectivity schemes}
\label{sec:nn-connectivity}

\subsection{In biological systems}
\begin{figure}
  \centering
  \begin{subfigure}{0.19\textwidth}
    \includegraphics[width=\textwidth]{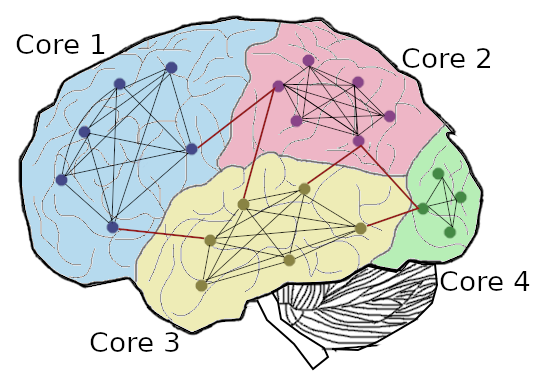}
    \subcaption{}
    \label{fig:small-world-network-brain}
  \end{subfigure}
  \begin{subfigure}{0.29\textwidth}
    \includegraphics[width=\textwidth]{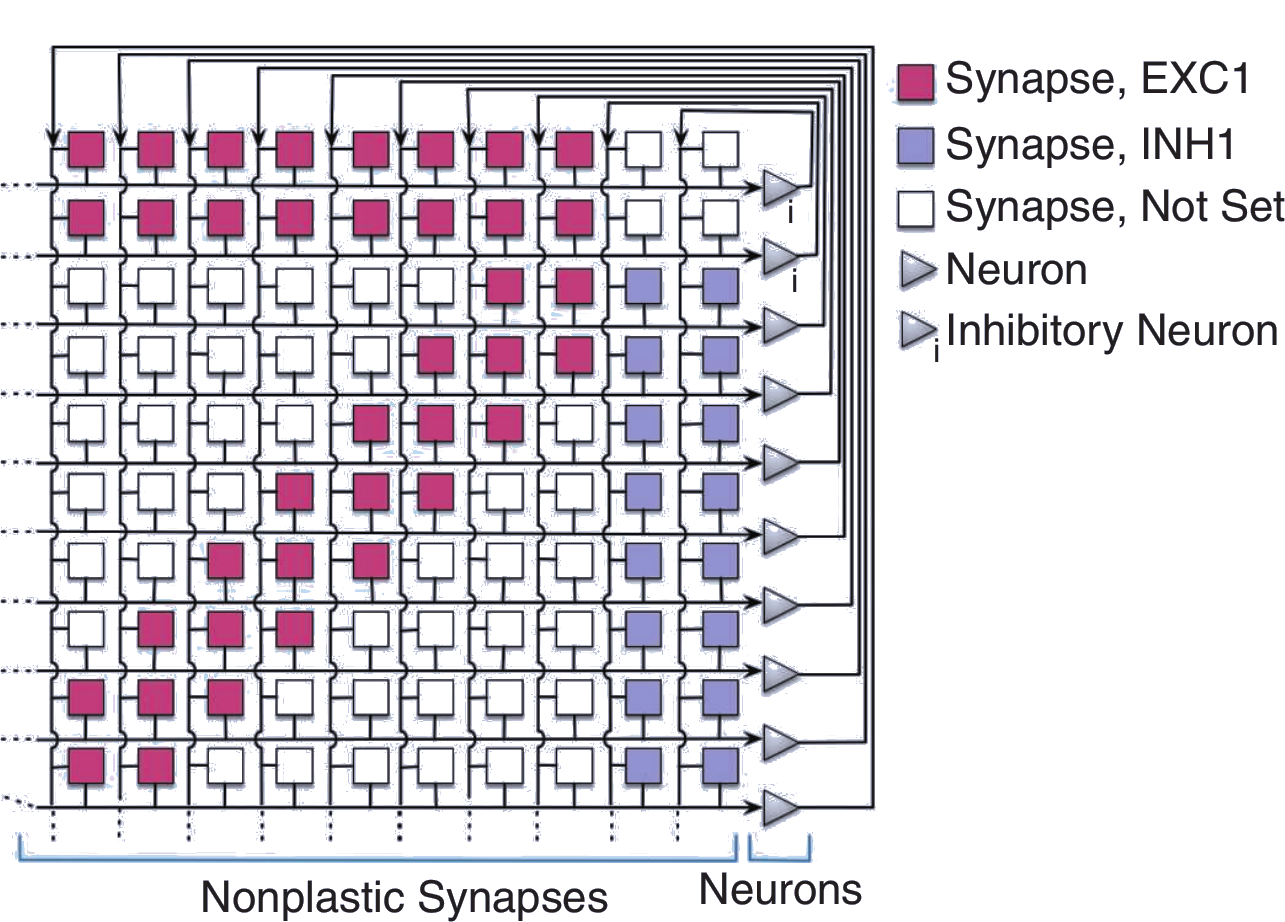}
    \subcaption{}
    \label{fig:wta-matrix}
  \end{subfigure}
  \caption{Small-world networks in brains and neuromorphic chips.
(\subref{fig:small-world-network-brain}) Schematic example of the patterns of connections found within and across different brain areas, which are highly correlated with brain functions~\cite{Lynn_Bassett19}.
Biological neural networks are often highly recurrent and have dense connections among nearby neurons and sparse connections to specific/far-away neurons, following an exponential decay in the number of connections with increasing distance.
(\subref{fig:wta-matrix}) Example of a ``Winner-Take-All'' (WTA) network implemented on a neuromorphic processor (from~\cite{Indiveri_Sandamirskaya19}).
Blue squares represent inhibitory synapses with negative weights, and red squares represent excitatory synapses with positive weights.}
  \label{fig:small-world-networks}
\end{figure}

In animal brains, computation and other functions emerge from the interaction of neural areas.
Brain networks have short path length, high clustering, and a modular community structure~\cite{Bullmore_Sporns09}.
They express modular, \emph{small-world}, heavy-tailed characteristics.
In small-world networks, most edges form small, densely connected clusters and the others maintain connections between these clusters (Fig.~\ref{fig:small-world-network-brain}).
This mixture of local clusters and global interaction generates a structure that provides function and integration in the brain that can support a wide range of complex computation, cognition and behavior~\cite{Bullmore_Sporns09}. 
By restricting the types of \acp{SNN} that can be implemented in neuromorphic processors to small-world networks, we can dramatically reduce the memory required to specify the routing/connectivity schemes while still supporting a wide range of computations for solving pattern recognition and signal processing tasks, e.g.~\cite{Donati_etal18b, Donati_etal19, Risi_etal21, Krause_etal21, Kreiser_etal20}.

\subsection{Routing schemes in neuromorphic hardware}

Multi-core neuromorphic processors usually use \ac{NOC} designs for managing the communication of neurons between cores.
Different neuromorphic chips adopt different \ac{NOC} architectures, according to the application.
Mesh architectures~\cite{Davies_etal18} represent an easy way to build large-scale systems, however, when the \ac{NOC} size increases, the required hardware area increases considerably which reduces the system scalability.
In flattened butterfly architectures~\cite{Chen_etal19}, neuron cores belonging to the same row and column can communicate directly, with lower routing latency, but this architecture also brings the disadvantage of large area cost and poor multi-casting support.
In~\cite{Park_etal16}, the authors proposed a hierarchical architecture that overcomes some of these disadvantages, by using off-chip \ac{DRAM} to store the routing lookup table, which significantly increases power consumption.

Current methods for saving power adopt in- or near-memory computing strategies.
However, when on-chip memory is used to store configurable neuron connections, the required hardware area increases proportionally with the number of neurons and synapses.
For example,~\cite{Moradi_etal18} uses on-chip hierarchical routing with a combination of point-to-point source-address routing and multicast destination-addresses to reduce memory usage, and still, the memory used takes around 80\% of the chip area.

\subsection{Network placement on neuromorphic hardware}

The \ac{NOC} and routing scheme define the source and target memory structures, thus setting constraints and restrictions on placing a network on it.
Placing an \ac{SNN} onto neuromorphic hardware is a mandatory step, needed to exploit the advantages of the hardware~\cite{Mysore_etal22,Balaji_etal20}, and each type of hardware has its own set of tools to make it appealing to \ac{SNN} developers.
Two main approaches are used to offer such a set of tools: platform-based design and hardware-software co-design.
The approaches proposed in~\cite{Amir_etal13,Galluppi_etal12a,Lin_etal18} are platform-based designs, where the development of the hardware is independent of its software, allowing exploration of alternative solutions, in a more general setting.
In this work we use the second approach, where the memory minimization strategies validated in software lead to routing circuit specifications for new neuromorphic chip designs.
The contributions of this paper are two-fold: i) a novel architecture designed to support small-world networks, and ii) a new placement algorithm to provide specifications for new hardware designs.
When developing the placement algorithm we take into account requirements derived from hardware design choices proposed by the chip designers, which define constraints on the algorithm.
In this way, software and hardware are optimized together.

\section{Hardware-software co-design strategy}
\label{sec:hardware-software-codesign}

By limiting the topology of the networks to be small-world networks, we can minimize memory requirements by reducing the address space (i.e., the number of bits and hence chip area) required to map the (many) connections between nearby neurons, and allocate more bits for larger address space domains used by the sparse long-range connections.
We took as ``canonical'' examples of \ac{WTA} networks, as shown in Fig.~\ref{fig:wta-matrix}, networks that have small-world connectivity matrices.
And, to generalize to other types of small-world networks, while minimizing memory usage, we propose a new heuristic for a hardware-aware \emph{neuromorphic compiler}.

\subsection{Hierarchical routing model}
\label{sec:hier-rout-model}

\begin{figure}
  \centering
  \includegraphics[width=0.25\textwidth]{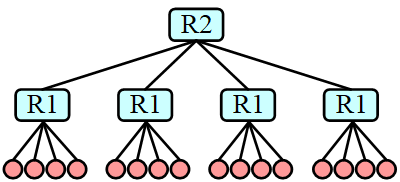}
  \caption{Hierarchical routing scheme: example with three levels of routers.
   R0 (not shown) connects neurons inside the same core (each core is shown as a red circle), R1 manages connections between $n$ local cores ($n=4$ in this example), and R2 connects $n$ R1 routers.
  With this routing scheme we can reduce dramatically the number of bits (memory) necessary to specify a small-world network, even with a large fan-in.}
  \label{fig:routing}
\end{figure}

Figure~\ref{fig:routing} shows our hierarchical routing scheme designed to support small-world network connectivity with on-chip memory.
Since we have densely connected clusters, there is no specificity within the cores in our architecture: independently of which neuron is sending a spike, all other neurons inside the same core will receive it.
Following the exponential decay of connections with distance observed in biology, we assume that the number of connections required between cores is dependent on the distance between them, and we associate physical distance with the levels in the router hierarchy: e.g., all cores that can be reached via an R1 router level are at distance 1.
Each neuron in a core reached through an R1 router can receive spikes from half of the neurons belonging to the source cores. Similarly, neurons reached through an R2 router will accept spikes from a fourth of the neurons of the source cores. As the distance between cores increases, fewer connections are made, thus, there is no need to allow connectivity between all neurons in different cores.
This allows us to reduce the connectivity address space and thus reduce the overall memory required to specify the source population address for each neuron.

These connections have a constrained address space, i.e., since there is an upper bound on the router to which a spike can be sent, it is not necessary to take into account all of the neurons on the whole chip.
In a model going up to the R2 level, each neuron needs memory to store 10 bits in total, allowing a fan-in from up to half of the neurons in the cores that can be reached through the R1 level plus a fourth of the neurons in each of the cores that can be reached through R2.
To support this reduction in address space we compute the routing distance by combining the use of computing logic and memory in each router module and update the distance information in the spike packet as it traverses the router, thus reducing the address space to the bare minimum needed by the local cluster.

\subsection{Placement algorithm}
\label{sec:placement-algorithm}

We present a placement algorithm (Alg.~\ref{alg:placement}) that optimally maps \ac{SNN} models that follow a small-world structure onto neuromorphic hardware architectures that implement the specificity and distance-based connectivity constraints described in Section~\ref{sec:hier-rout-model}.

\begin{algorithm}
\floatname{algorithm}{Alg.}
\caption{High-level description of the placement algorithm}\label{alg:placement}
\begin{algorithmic}
\Require $G=(V,E)$, where $V$ is the set of neurons and $E$ is the set of paired neurons.
\State find cliques on $G$
\For{each clique}
\State Add clique to a new core
\EndFor
\State $n = $ number of neurons per core
\State $e(i,j) = $ number of connections core $i$ receives from core $j$.

\State initialize matrix $dist[i][j]$ with zeros
\For{ each pair of cores $i$ and $j$}:
\If{$e(i,j) == 0$}
$\ \mathit{dist}[i][j] = -1$
\Else
$\ \mathit{dist}[i][j] = \mathit{floor}((n/e(i,j)))+1$
\EndIf
\EndFor
\State sort $\mathit{dist}[i][j]$
\For{each pair of cores $i$ and $j$ in $\mathit{dist}$}
\If{neuron in $i$ has synapse available}
\State connection is placed
\Else
\State connection is flagged
\EndIf
\EndFor

\end{algorithmic}
\end{algorithm}

To place neurons in cores, we first find \emph{cliques} in the network graph.
Each clique is a group of densely connected neurons, and they will be placed in different cores.
(If there are cliques larger than the core size, they can be subdivided.)
This first step gives us the number of cores needed to place the network.
Then we calculate the distance between two cores $i$ and $j$, as a function of the number of connections that they share.
We define the distance between the core and itself ($i=j$) as zero, and between two cores that do not share connections as --1.
Note that our distance definition is a \emph{quasi-metric} that can be non-symmetric, i.e., the distance from core $i$ to $j$ might not be the same as from $j$ to $i$.
The maximum distance in the network indicates the maximum router level we need to map the network.

Having determined the number of cores and distances between cores, we can finally place connections.
We start the connection placement from the closest pair of cores.
The closer cores are to each other, the more connections they share.
We allocate the nearby connections first because they are more numerous than the further-away ones.
If a connection can not be added, it is flagged as unplaced.
Unplaced neurons and connections can be added in a second loop, given the availability of extra cores.

\section{Results}

\begin{figure}
  \centering
  \includegraphics[width=0.25\textwidth]{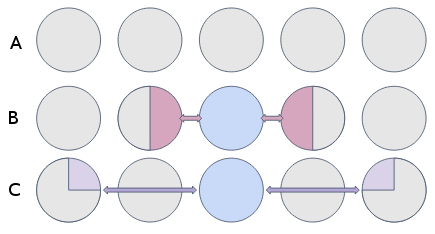}
  \caption{Canonical networks generated to match a hypothetical neuromorphic processor.
    Row \textsc{\textsf{a}} shows five populations.
	We create populations with the same number of neurons as in a core, all-to-all connected.
    For simplicity, we place our populations on a 1D-line defining distances (and thus the connections between populations).
    In \textsc{\textsf{b}} we assign connections between cores at distance 1.
    These cores will share connections with half of the number of neurons in a nearby core.
    In \textsc{\textsf{c}} we assign connections between cores at distance 2.
    They share connections with a fourth of the number of neurons in a core at that distance.
    This process is repeated until there are no more cores or they are so far away that no connection is created.
    Only the procedure for the central core is shown in the figure, but this process is applied to all cores.
  }
  \label{fig:canonical-network}
\end{figure}

To validate our placement algorithm we tested it with canonical networks generated to match a hypothetical neuromorphic processor comprising cores of 16 neurons.
The canonical network thus have populations of 16 neurons, where each population is all-to-all connected, and the number of connections between populations drops off with the distance between the cores, as depicted in Fig.~\ref{fig:canonical-network}.

By using a canonical network that fully matches the hardware structure considered, we define a \emph{ground truth} (GT) placement.
As our algorithm maps the canonical network to this hardware perfectly, we can verify that the proposed heuristic works as expected.

To evaluate how the algorithm performs in non-optimal conditions, we performed two sets of tests.
Two canonical \ac{WTA} networks were considered, based on hardware with 16 neurons per core, and either 7 or 70 cores (112 or 1120 neurons respectively).
First, we tested it by using perturbed networks that deviate from the canonical one by removing a percentage of nodes (1\%, 10\% and 25\%).
With this experiment we evaluate how much deviations from the canonical network affect the placement algorithm.
Our algorithm still finds solutions that are very close to the GT, for small deviations (1\% and 10\%).
Larger deviations (25\%) produce solutions that are close to the GT only in large networks ($\sim$1k neurons).

\begin{figure}
  \centering
  \begin{minipage}{0.24\textwidth}
    \centering
    \includegraphics[width=\linewidth]{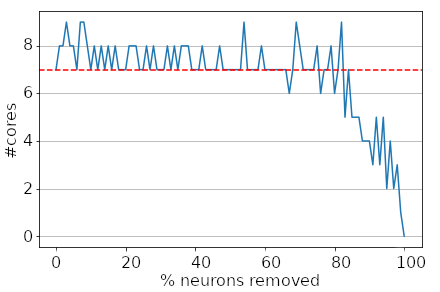}
    \subcaption{}
  \end{minipage}
  \begin{minipage}{0.24\textwidth}
    \centering
    \includegraphics[width=\linewidth]{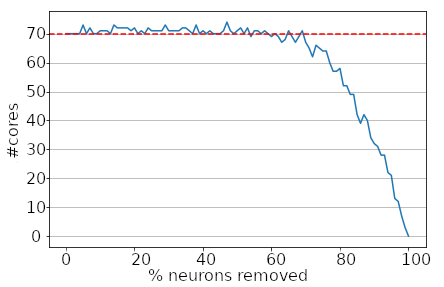}
    \subcaption{}
  \end{minipage}
	\caption{Placement of SNN network models onto the resource constrained neuromorphic hardware.
	Two canonical \ac{WTA} networks were considered, based on hardware with 16 neurons per core, and either 7 or 70 cores.
	The red dashed line marks the canonical network placement.
	For each one of the canonical networks, we remove an increasing number of neurons and place the perturbed network.
	(a) shows the result for the $\sim$100-neuron network, and (b) for the $\sim$1k-neuron network.
	Note that for the perturbed networks, there is some fluctuation in the number of cores needed, however it stays close to the ideal case.
}
	\label{fig:preliminary-results}
\end{figure}

Secondly, we tested the placement algorithm by removing an increasing number of nodes (from $1$ to all the nodes in the network).
The results of this test are shown in Figure~\ref{fig:preliminary-results}.

For all perturbed networks and network sizes considered, the performance of the placement algorithm is close to the GT performance, indicating an optimal use of the limited resources on the neuromorphic hardware.

\subsection{Memory comparison}

\begin{figure}
  \centering
  \includegraphics[width=0.3\textwidth]{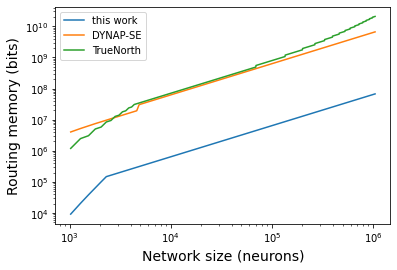}
  \caption{Memory scaling comparison between TrueNorth~\cite{Merolla_etal14a} and \ac{DYNAP}~\cite{Moradi_etal18} architectures and this work.
  For comparison, we considered a core with 256 neurons in all architectures.
  The number of bits used by the three architectures is plotted as a
function of the network size (log scale).
For TrueNorth and \ac{DYNAP}, the memory increases linearly with increasing fan-in/fan-out,
with TrueNorth performing better than \ac{DYNAP} while the fan-in fits the cores.
In our proposed architecture, the increase in fan-in/fan-out demands more cores but the memory required increases at a slowly, since all the cores at the same level need a fixed number of bits to be addressed.
}
  \label{fig:memory-comparison}
\end{figure}

In Fig.~\ref{fig:memory-comparison} we compare our work with TrueNorth~\cite{Merolla_etal14a} and \ac{DYNAP}~\cite{Moradi_etal18} architectures.
In our design, the number of bits required to fit a network does not increase with the neuron fan-in/fan-out.
We can increase the fan-in by adding more cores, so the increase in memory is linear in the number of cores.
TrueNorth architecture has a fixed fan-in per neuron, and with an increase in fan-in/fan-out we need to recruit relay neurons from other cores.
This starts to be costly for large networks in which neurons have a large fan-in.
Indeed, any architecture with a fixed fan-in per neuron will not scale well due to the requirement to resort to relay neurons~\cite{Rao_etal22}.
Also in the \ac{DYNAP} architecture the fan-in is fixed. But its mixed source/destination addressing scheme mitigates the number of intermediate nodes required.
Our analysis shows that our canonical network with a million neurons requires $\sim$67\,Mbit if implemented with our scheme, about 98$\times$ more on \ac{DYNAP} ($\sim$6591\,Mbit) and about 307$\times$ more on TrueNorth ($\sim$20649\,Mbit).

\section{Conclusion}
The development of domain-specific neuromorphic hardware can help to advance \ac{AI} for edge-computing tasks, and the optimization of memory resource allocation paves the way to building large-scale neuromorphic computing systems.
In this work, we present a hardware-software co-design approach, where brain-like small-world networks are used to inspire simultaneously our routing scheme and placement algorithm.

Our co-design approach reduces the memory necessary to place and route networks that follow a small-world structure while not limiting the possible applications.
Additionally, our placement algorithm can find optimal solutions for networks that follow our canonical design and can place deviations from them without diverging too much from the ideal case.

The simultaneous design of a place and route scheme is allowing us to design a new multi-core \ac{SNN} chip able to handle larger networks with a minimum of memory consumption, and thus smaller area.
%

\bibliographystyle{ieeetr}
\bibliography{biblioncs}

\end{document}